%% file: main.tex
\documentclass[twoside]{article}

\usepackage[accepted]{aistats2023}
\usepackage[utf8]{inputenc} 
\usepackage[T1]{fontenc}    
\usepackage{hyperref}       %
\hypersetup{
    colorlinks=true,
    linkcolor=red,
    citecolor=blue,      
    }

\usepackage{url}            
\usepackage{booktabs}       
\usepackage{amsfonts}       
\usepackage{nicefrac}       
\usepackage{microtype}      
\usepackage{xcolor}         
\usepackage{graphicx}
\usepackage{subfigure}
\usepackage{amsmath, xparse}
\usepackage{siunitx}
\usepackage{float}
\usepackage{multirow}
\usepackage{marvosym}

\usepackage[authoryear]{natbib}

\begin{document}

%

%
\runningauthor{Mingshan Sun, Ye Zheng, Tianpeng Bao, Jianqiu Chen, et al.}

\twocolumn[
\aistatstitle{Uni6Dv2: Noise Elimination for 6D Pose Estimation}

\aistatsauthor{Mingshan Sun \And  Ye Zheng\textsuperscript{\Letter} \And  Tianpeng Bao \And Jianqiu Chen}

\aistatsaddress{SenseTime Research \And  JD.com, Inc \And SenseTime Research \And Harbin Institute of \\ Technology, Shenzhen }

\aistatsauthor{Guoqiang Jin \And Liwei Wu \And  Rui Zhao \And  Xiaoke Jiang }

\aistatsaddress{ SenseTime Research \And  SenseTime Research \And SenseTime Research \And International Digital \\ Economy Academy (IDEA) }
]


\input{sec/0abstract}
\input{sec/1intro}

\input{sec/2related}
\input{sec/3approach}
\input{sec/4exp_overview}
\input{sec/5exp_ablation}

\input{sec/6conclusion}

\bibliography{ref}
\input{sec/supplement}

\end{document}

%% file: sec/0abstract.tex
\begin{abstract}
Uni6D is the first 6D pose estimation approach to employ a unified backbone network to extract features from both RGB and depth images. We discover that the principal reasons of Uni6D performance limitations are Instance-Outside and Instance-Inside noise. Uni6D's simple pipeline design inherently introduces Instance-Outside noise from background pixels in the receptive field, while ignoring Instance-Inside noise in the input depth data. In this paper, we propose a two-step denoising approach for dealing with the aforementioned noise in Uni6D. To reduce noise from non-instance regions, an instance segmentation network is utilized in the first step to crop and mask the instance. A lightweight depth denoising module is proposed in the second step to calibrate the depth feature before feeding it into the pose regression network. Extensive experiments show that our Uni6Dv2 reliably and robustly eliminates noise, outperforming Uni6D without sacrificing too much inference efficiency. It also reduces the need for annotated real data that requires costly labeling.
\end{abstract}

%% file: sec/1intro.tex
\section{INTRODUCTION}
\label{Introduction}

\begin{figure}[htbp]
	\centering
	\subfigure[]{\includegraphics[width=0.85\linewidth]{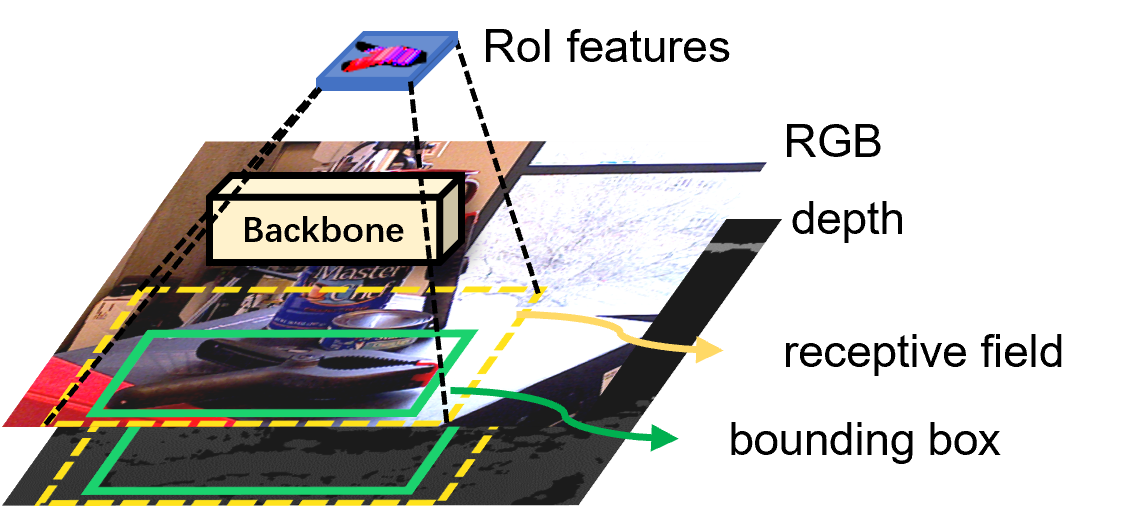}%
	    \label{fig:fig1_outside}
		}
	\subfigure[]{\includegraphics[width=0.85\linewidth]{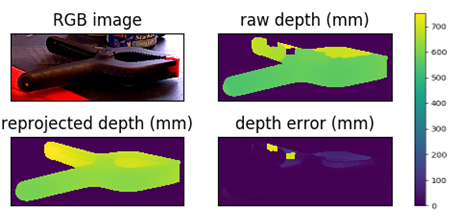}%
	    \label{fig:fig1_inside}
		}\\\vspace{-0.4cm}
	\subfigure[]{\includegraphics[width=0.95\linewidth]{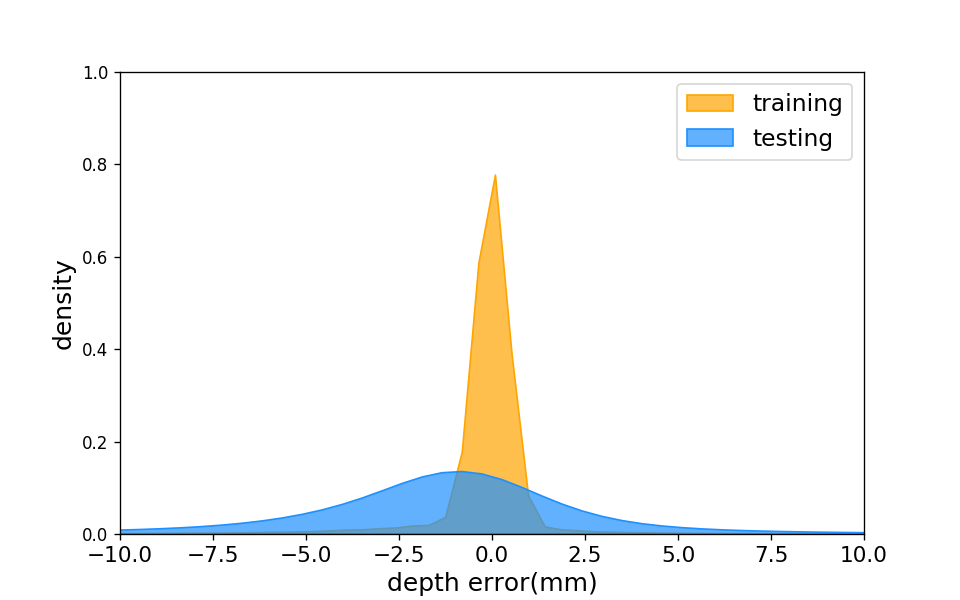}%
	    \label{fig:fig1_1}
		}
	\subfigure[]{\includegraphics[width=0.85\linewidth]{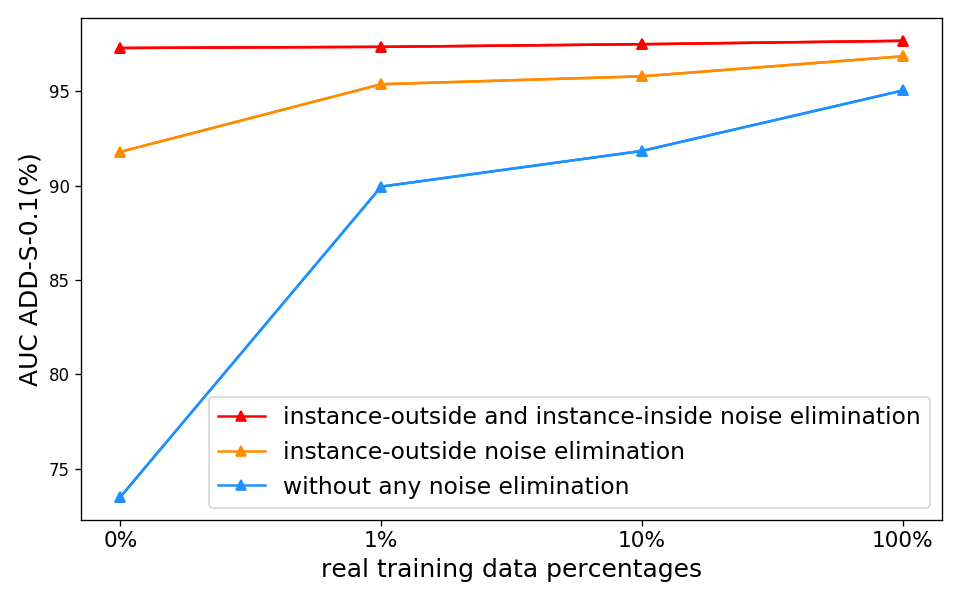}%
		\label{fig:fig1_2}}

    \caption{Illustration of noise problem in 6D pose estimation.
     (a) Examples of instance-outside noise, which comes from background pixels outside the target instances. 
     (b) Examples of instance-inside noise, which comes from the unreliable depth data.
     (c) Quantitative results of instance-inside noise according to the depth reprojection error statistics. 
     (d) Uni6D results under various real data percentages and noise elimination methods.
     Eliminating the noise improves performance and reduces the need for real data in training set.
     }
	\label{fig:hor_2figs_1cap}
\end{figure}

6D pose estimation is critical for upcoming applications including intelligent robotic grasping\Citep{collet2011moped,tremblay2018deep}, autonomous driving\Citep{geiger2012we,xu2018pointfusion,chen2017multi}, and augmented reality\Citep{marchand2015pose}.
The basic purpose of 6D pose estimation is to identify an object's 6D pose, including its location and orientation.
As RGB-D sensors become more possible and inexpensive, they have the potential to augment the typical RGB image with depth information on a per-pixel basis and offer direct geometry information, making it a more appealing data source for 6D pose estimation.
The only fly in the ointment is that it inevitably introduces physical noise when acquiring depth information\Citep{zhang2012microsoft, mallick2014characterizations,zhang2018deep, sweeney2019supervised, ji2021calibrated}.

Given the heterogeneity of RGB and depth data, previous state-of-the-art methods\Citep{densefusion,pvn3d,ffb6d} typically handle them independently with two backbone networks separately, in which, 2D CNN is used for RGB data and PointNet\Citep{pointnet} or PointNet++\Citep{pointnet++} for depth data.
Recently, a new approach called Uni6D\Citep{uni6d} was developed to overcome the "projection breakdown"\Citep{uni6d} problem by inserting extra UV information, i.e., coordinates of each pixel, into the RGB-D input. It uses basic framework of Mask R-CNN\citep{maskrcnn}, and especially leverages the unified CNN backbone to extract features from RGB and depth images possible, resulting in an efficient and straightforward realization pipeline.
However, Uni6D ignores the potential pitfalls hidden in its straightforward pipeline and the input depth data, severely limiting its performance.
%

We find that the main sources of Uni6D's potential pitfalls are two types of noise: instance-outside noise introduced by the RoI-based 6D pose estimation methodology, and instance-inside noise from the unreliable depth data.
As shown in Fig.~\ref{fig:fig1_outside}, the instance-outside noise comes from the background pixels outside the target instances, and Uni6D introduced it in the pose regression with its detection and pose regression pipeline. Since the deep features of CNN have corresponding receptive fields, the information out of the instance will also be included in the RoI features obtained by the RPN and RoI Align. Because the feature outside the instance isn't useful for pose estimation, this background information can even interference the pose regression. For the instance-inside noise, it is introduced by depth sensor during capturing depth information, as exampled in Fig.~\ref{fig:fig1_inside}.

We construct several corresponding experiments to quantify the above noise and its impact. We provide the statistic of instance-inside noise of the YCB-Video\Citep{posecnn} dataset in Fig.~\ref{fig:fig1_1}, by calculating the depth deviation according to depth reprojection error, which indicates that there is depth noise in the dataset. Then, in Fig.~\ref{fig:fig1_2}, we investigate the effects of performance by incrementally removing these kinds of noise. To filter out instance-outside noise, we crop and mask the RoI of each instance using the ground truth bounding box and mask. To reduce instance-inside noise, we use the depth calculated by reprojection as the ground truth.
We can observe that removing these two types of noise gradually can significantly improve the accuracy on the YCB-Video dataset and reduce the need for real data to train the model.
According to the preceding experimental results, we believe that each type of noise will have an effect on the model's performance. Worse still, limited by the high cost of annotation, existing datasets such as YCB-Video\Citep{posecnn} and LineMOD\Citep{hinterstoisser2011multimodal} include a lot of synthetic data without depth noise in training set and put the real data with depth noise in test set. This train-test gap, as shown in Fig.~\ref{fig:fig1_1}, accentuates the unfavorable effect of depth noise. 

%
To this end, we propose Uni6Dv2, a simple yet effective two-step denoising 6D posture estimation method. Uni6Dv2 predicts the instance segmentation mask of each instance from the RGB-D image in order to filter out extraneous non-instance pixels in the first step. A depth denoising module is then utilized in the second step to correct the depth feature with the relevant depth normal and XY before feeding it into the succeeding pose regression network.
%
%

%

Overall, the main contributions of this work are as follows:
\begin{itemize}
    \item We uncover different types potential noise that severely limit the performance of RoI-based 6D pose estimation methods, and categorize them as instance-outside noise and instance-inside noise;
    \item We propose a two-step denoising pipeline to address the noise problem in the original Uni6D, which uses the instance segmentation network to filter out the instance-outside noise in the first step and a depth denoising module to handle instance-inside noise in the second step; 
    \item
     The proposed approach achieves 96.8$\%$ on the AUC ADD-S metric for the YCB-Video dataset, advancing the state-of-the-art method FFB6D by 0.2$\%$ .
     In particular, our approach significantly outperformed Uni6D with a margin of 1.6$\%$ given 100$\%$ real training data and a margin of 20.1$\%$ given 0$\%$ real training data.
\end{itemize}

%% file: sec/2related.tex
\section{RELATED WORK}
%

\subsection{6D Pose Estimation}
In accordance with the way of 6D pose parameter estimation, we can classify 6D pose estimation methods into two categories: keypoint-based and RoI-based methods.  

\textbf{Keypoint-based Methods.}  
The 6D pose parameters in keypoint-based methods are estimated using a PnP algorithm that matches the predicted keypoints with the target keypoints.
Previous methods typically predict 2D keypoints through direct regression\Citep{rad2017bb8, tekin2018real,hu2019segmentation} or heatmaps\Citep{kendall2015posenet, newell2016stacked,oberweger2018making}.
%
Considering robustness in truncated and
occluded scenes, PVNet\Citep{pvnet} proposes a voting network to obtain the dense prediction of 2D keypoints of objects and utilizes an uncertainty-driven PnP algorithm to improve performance.
Recent methods\Citep{pvn3d, ffb6d} extend keypoint detection to 3D space by predicting each 3D point semantic label and offsets to pre-defined keypoints. 
To calculate keypoints in the camera coordinates system and distinguish different objects, an iterative voting mechanism is adopted to vote for the best prediction of keypoints based on the offsets and semantic labels. 
The final 6D pose parameters are predicted by an iterative least-squares regression algorithm, fitting predicted 3D keypoints with corresponding pre-defined 3D target keypoints.
However, the iterative voting and regression operations in these methods are time-consuming and heavy in practical applications.

%
\textbf{RoI-based Methods.} To estimate 6D pose of a single instance from the image with multiple instances, RoI-based methods crop each instance by the candidate RoI prediction and then feed each RoI region image patch or feature into the pose regression network to estimate the 6D pose directly.
PoseCNN\Citep{posecnn} utilizes two RoI pooling layers\Citep{fastrcnn} to extract the corresponding visual feature to regress the quaternion representing rotations. 
%
Based on the RoI feature from RGB images, following methods\Citep{silhonet, mcn, gdrnet} improve the computation method of rotation as well as translation and introduce extra class and geometric priors.
However, the insufficiency of geometry information limits the performance of these methods.
To make use of geometry information, DenseFusion\Citep{densefusion} and ES6D\Citep{es6d} crop the RoI region from both RGB and depth images and then concatenate them as the input of pose regression network.
In practice, the crop and RoI-pooling operations in these CNN pipelines lead to spatial transformation, which breaks the 2D-3D projection equation.
Uni6D\Citep{uni6d} is the first to expose the "projection breakdown" problem and solve it by adding extra UV information into the RGB-D input.
Hence, it can adopt a single CNN backbone to process heterogeneous data sources.
However, the performance of the method degrades when there is noise in the candidate RoI regions. In the following subsection, we conduct an in-depth 
 and systematic analysis of the denoising operations in 6D pose estimation, which can alleviate the interference from noise.

\subsection{Denoising Operations in 6D Pose Estimation}

Through the investigation of the existing 6D pose estimation methods, the potential denoising operations mainly include two types: vote-based denoising and mask-based denoising.

\textbf{Vote-based Denoising.}
The voting mechanism iteratively selects the centroid or keypoints based on the dense pixel-level prediction.
Pixels that contribute to the voting process are devoted to predicting the final pose estimation.
Conversely, the rest pixels viewed as inaccurate predictions are not involved in the final decision, thus performing denoising in an implicit way.
PoseCNN\citep{posecnn} estimates 2D object centroid by the vote of the pixel-wise prediction for the object center direction vector, instead of directly regressing the coordinates.
The voting mechanism filters out the predictions of unvoted pixels, alleviating the interference caused by noise.
For keypoint prediction, PVNet\Citep{pvnet} adopts the same vote mechanism to estimate the location of 2D keypoints by the pixel-wise vector-field prediction.   
To extend keypoints detection from 2D to 3D, PVN3D\Citep{ pvn3d} and FFB6D\Citep{ffb6d} employ a voting mechanism to predict 3D keypoints and estimate the 6D pose by fitting the predicted keypoints to their pre-defined counterparts iteratively.
The fitting algorithm is able to suppress noise in the instance-inside points.
However, the iterative voting and fitting process used for denoising are inefficient and heavy, accounting for a significant portion of the total frame processing time.

\textbf{Mask-based Denoising.}
Prior works use a mask operation to remove noise from the instance-outside regions that can interfere with the target instance feature representation and affect the 6D pose regression. 
%
%
This is typically accomplished using a two-stage approach\Citep{posecnn, silhonet, mcn,self6d,es6d},
in which an instance segmentation network is first used to generate semantic segmentation masks and bounding boxes for each target instance. In the second stage, a pose estimation network is trained on the cropped or masked RoI features to predict the 6D pose.

%
%
%
%
To further mitigate the interference from the instance-outside noise, recent work ES6D\citep{es6d} estimates 6D pose based solely on the point with the highest confidence score.
However, the instance-outside region is still present in the input of the network for estimate 6D pose.
%
To comprehensively address the noise issue, we systematically analyze the source of noise and propose a novel two-step pipeline with a lightweight depth denoising module.

%% file: sec/3approach.tex
\section{APPROACH}
%
\label{headings}
Our work can be viewed as an extension of Uni6D\Citep{uni6d} with a novel two-step denoising pipeline. We first review Uni6D (Sec.~\ref{sec:review_uni6d}), and then give the details of the proposed method, including the denoising processes of instance-outside noise and instance-inside noise (Sec.~\ref{sec:method}). Finally, the details of loss function are provided (Sec.~\ref{sec:loss}).

\subsection{Review of Uni6D: a Unified CNN Framework for 6D Pose Estimation}
\label{sec:review_uni6d}
Uni6D addressed the "projection breakdown" problem, which arises from the CNN and spatial transformations, by adding extra visible point coordinates in both the image plane and 3D space (Plain UV, XY) with depth normal(NRM) into RGB-D images. This enables simultaneous feature extraction of RGB and depth data using a unified network. 
It took the Mask R-CNN structure and extended it with an extra RT head for estimating rotation and translation parameters, and an abc head regressing visible points $(a,b,c)$ from the CAD model to directly estimate the 6D pose. It achieved impressive results on existing 6D pose estimation benchmarks. However, it suffered from instance-outside noise caused by background pixels and ignored instance-inside noise in the input depth data. In the following section, we will present our two-step denoising method to tackle these challenges.


\begin{figure*}[t]

    \centering
    \includegraphics[width=0.8\textwidth]{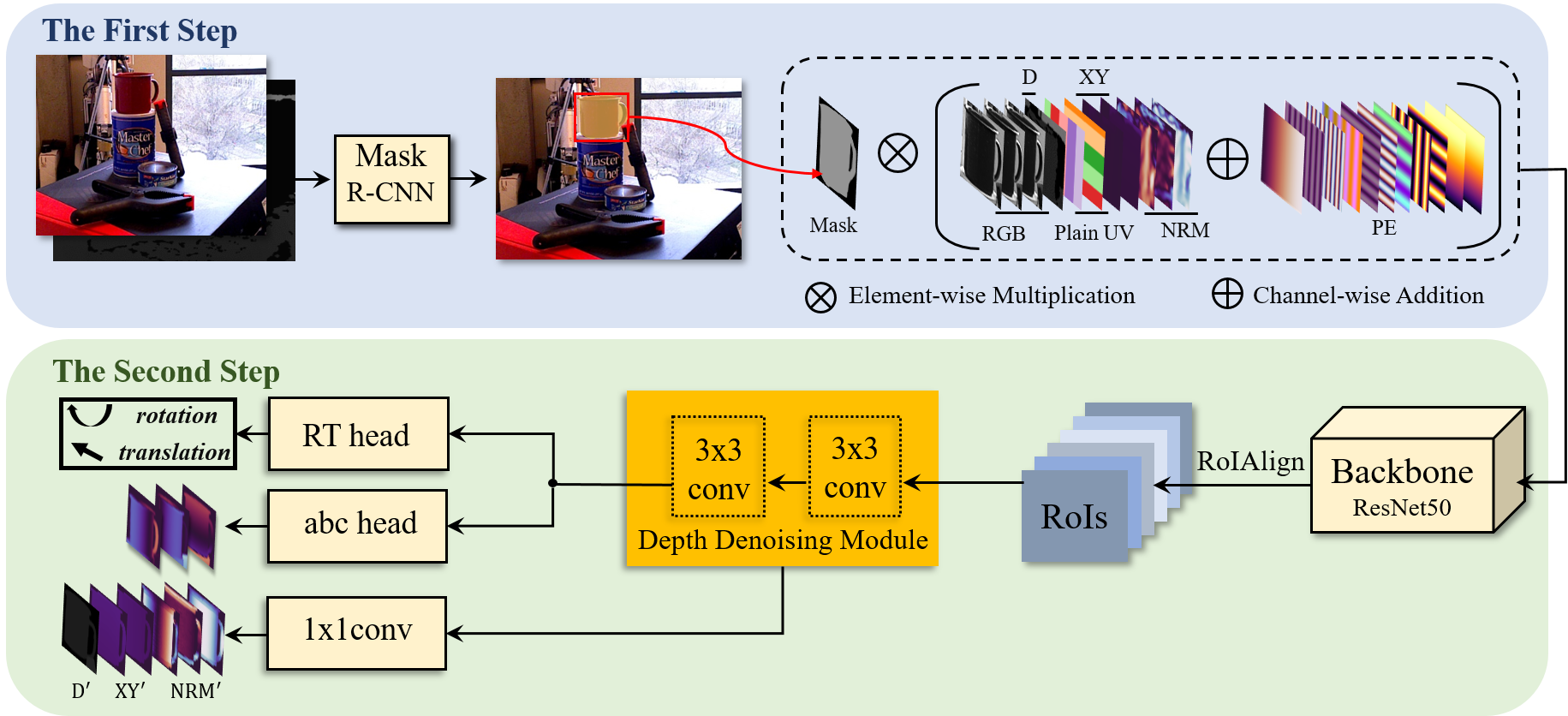}

    \caption{Architecture overview of Uni6Dv2. Uni6Dv2 consists of two denoising steps: the first step removes instance-outside noise by adopting the Mask R-CNN to crop and mask the information about the instance, while the second step removes the instance-inside noise by utilizing a depth denoising module and a depth information estimation auxiliary task to calibrate the depth feature before the 6D pose regression. The auxiliary supervision labels $D^{'}$, $XY^{'}$ and $NRM^{'}$ are calculated by projecting the CAD model with the rotation and translation ground truth.}
    \label{fig:framework}
\end{figure*}

\subsection{Comprehensive Denoising Pipeline}
\label{sec:method}
To effectively suppress the instance-outside and instance-inside noise, we propose a novel two-step denoising pipeline called Uni6Dv2, as illustrated in Fig.~\ref{fig:framework}. Our method builds upon the basic methodology of Uni6D, which involves adding various UV information to input RGB-D data and regressing 6D pose directly. 
The instance-outside noise is handled in the first denoising step. For an input RGB-D image, we perform instance segmentation with Mask R-CNN to filter out the instance-outside pixels in RGB, depth, Plain UV, XY, NRM and corresponding position encoding(PE).
The above data is then fed into the second step, where we introduce a depth denoising module and a depth reconstruction task to calibrate the features extracted from the output of the first step. This second step facilitates more accurate 6D pose regression by subsequent networks. We use Uni6D's RT head to regress the 6D pose directly, while regressing corresponding coordinates in CAD model by abc head as the auxiliary training branch.

\textbf{Instance-outside Noise Elimination.} 
The most intuitive and efficient choice to filter out the instance-outside noise is directly discarding non-instance regions on the feature map. 
However, this feature-level operation falls short of entirely removing the background feature noise, as the deep features in CNN have a receptive field includes the surrounding background information.
%
To comprehensively address this problem, instance-outside noise must be filtered out at the image level. 
%
We adopt Mask R-CNN to obtain the bounding boxes and segmentation results for all instances in the input RGB-D image, 
allowing us to crop RGB-D patches  based on these bounding boxes to eliminate the majority of the instance-outside noise. And the rest of instance-outside noise, which remains in the bounding box, is further removed by masking cropped patches with segmentation results.
Corresponding Plain UV, XY, and NRM and PE are also cropped and masked.
%

%
\textbf{Instance-inside Noise Elimination.} 
As shown in Fig.~\ref{fig:fig1_inside}, the instance-inside noise includes holes and numerical errors in the depth image, which limit the performance of the 6D pose estimation.
A widely used method to address this problem is to utilize the classical processing algorithm\Citep{ku2018defense} to fill the holes in the raw depth image before regressing the 6D pose. However, the numerical errors in the filled pixels still persist due to the non-parametric pre-processing method.
To handle the instance-inside noise more effectively, we propose a learnable parametric depth denoising module with a noiseless depth estimation task to calibrate the depth information in the features. As shown in the second step of Fig.~\ref{fig:framework}, features extracted from RoI Align are fed into our depth denoising module before being used to regress the 6D pose.
We also incorporate an auxiliary reconstruction task to help the depth denoising model converge better. It uses a $1\times 1$ convolutional layer to regress the noiseless depth, XY, and NRM simultaneously.
%
The labels of the noiseless depth, XY, and NRM are obtained by re-projection from the CAD model.
These re-projected labels can be calculated as follows:
\begin{equation} \label{equ:prj}
  \begin{bmatrix}
    x \\
    y \\
    d 

  \end{bmatrix}
    =
    R^{*} \times
  \begin{bmatrix}
    a  \\
    b  \\
    c 
  \end{bmatrix}
  + T^{*}.
 \end{equation}
For an input visible point $(a,b,c)$ in the CAD model, we apply the ground truth rotation matrix $R^{*}\in SO(3)$ and the translation matrix $T^{*}\in R^3$ to project it to the corresponding position $(x,y,d)$ in camera coordinate system and obtain the re-projected depth and XY.
This lightweight depth denoising module can effectively calibrate depth information in features without reducing inference efficiency significantly.

\subsection{Loss Function}
\label{sec:loss}
The loss function of the overall network consists of the loss functions from both steps.
Initially, the Mask R-CNN is trained with its original loss~\citep{maskrcnn}.
%
Following this, we freeze the parameters of the first step and train the network in the second step with the RT regression loss, the abc regression loss, and the novel depth denoising loss. 
The RT regression loss $\mathcal{L}_{rt}$ is deﬁned as:
\begin{equation}
    \mathcal{L}_{rt} = \frac{1}{m}\sum_{x\in\mathcal{O}}||(Rx+T) - (R^*x+T^*)||,
\end{equation}
where $O$ denotes the vertex set of object’s points from the 3D model, $R$ and $T$ are the rotation matrix and the translation vector, $^*$ denotes the ground truth and $m$ is the number of points in $O$. 
The abc regression loss is:
\begin{equation}
    \mathcal{L}_{abc} = |a-a^{*}| + |b-b^{*}| + |c-c^{*}|,
\end{equation}
where $(a,b,c)$ are the coordinates of visible points. 
The depth denoising loss is:
\begin{equation}
\begin{aligned}
{L}_{depth} = &|d-d^{*}| + |x-x^{*}| + |y-y^{*}|+\\&  |n_x-n_x^{*}| +
|n_y-n_y^{*}| + |n_z-n_z^{*}|,
\end{aligned}
\end{equation}
where $d$ is the depth, $(n_x,n_y,n_z)$ is the coordinate of depth normal calculated from depth, and $(x, y)$ is the visible point's coordinate in camera coordinate system. Finally, the overall loss function of the second step is:
\begin{equation}\label{equ:loss}
	\mathcal{L} = \lambda_0  \mathcal{L}_{rt} + \lambda_1  \mathcal{L}_{abc} + \lambda_2 \mathcal{L}_{depth},
\end{equation}
where $\lambda_0$, $\lambda_1$ and $\lambda_2$ are the weights for each loss.

%% file: sec/4exp_overview.tex
\section{EXPERIMENTS}
\label{others}
\subsection{Benchmark Datasets}
\label{sec:datasets}
We compare the proposed method with others on  three benchmark datasets. 

\textbf{YCB-Video}\Citep{calli2015ycb} contains 92 RGB-D videos with 21 YCB objects, which has depth noise and occlusions. Following previous work\Citep{posecnn,pvn3d,densefusion,uni6d}, we add synthetic images for training and the synthetic ratio is 83.17$\%$. We also apply the hole completion algorithm used in\Citep{pvn3d,uni6d} to improve depth images.

\textbf{LineMOD}\Citep{hinterstoisser2011multimodal} contains 13 videos of 13 low-textured objects. Following previous works\Citep{peng2019pvnet,posecnn,pvn3d,uni6d}, we also add synthetic images for training and the synthetic ratio is 99.71$\%$.

\textbf{Occlusion LineMOD}\Citep{brachmann2014learning} is selected from the LineMOD dataset, which has heavily occluded objects, making the scenes more challenging.

\subsection{Evaluation Metrics}

We use the average distance metrics ADD and ADD-S to evaluate our method, following\Citep{posecnn,densefusion,ffb6d,uni6d}. The ADD metric\Citep{hinterstoisser2012model} is calculated as follows:
\begin{equation}
    \begin{split}
        \textrm{ADD} = \frac{1}{m}\sum_{x\in\mathcal{O}}||(Rx+T) - (R^*x+T^*)||,
    \end{split}
\end{equation}
where $m$ is the total vertex number, $x$ indicates any vertex of the 3D model $\mathcal{O}$,  $[R, T]$ and $[R^*, T^*]$ are predicted pose and ground truth pose respectively. The ADD-S is similar to ADD, but is used to measure the performance of symmetric objects, which is based on the closest point distance: 
\begin{equation}
    \begin{split}
        \textrm{ADD-S} = \frac{1}{m}\sum_{x_{1}\in\mathcal{O}}\min_{x_{2}\in\mathcal{O}}||(Rx_{1}+T) - (R^*x_{2}+T^*)||.
    \end{split}
\end{equation}
For convenience, the ADD(S) metric is introduced as follows:
\begin{eqnarray}
\textrm{ADD(S)} =
\begin{cases}
\textrm{ADD}   & \mathcal{O}~is~asymmetric\\
\textrm{ADD-S} & \mathcal{O}~is~symmetric 
\end{cases}
\end{eqnarray}
where $\mathcal{O}$ is the CAD model.

For YCB-Video dataset, the area under the accuracy-threshold curve obtained by varying the distance threshold with a maximum threshold of 0.1 meters (AUC ADD-S or AUC ADD(S)) is reported, following previous works \Citep{posecnn,densefusion,pvn3d,ffb6d}. 
For LineMOD and Occlusion LineMOD datasets, the accuracy of distance less than 10\% of the objects' diameter (ACC ADD(S)-0.1d) is reported, following previous works \Citep{peng2019pvnet,ffb6d}


\begin{table*}[ht]
\begin{center}
\caption{Evaluation results on the YCB-Video dataset. Symmetric objects are denoted in bold.}
\vspace{.15in}
\label{tab:ycb}
\resizebox{0.99\linewidth}{!}{
\begin{tabular}{lcccccccccccc}
    \toprule
 & \multicolumn{2}{c}{PoseCNN\Citep{posecnn}}  & \multicolumn{2}{c}{DenseFusion\Citep{densefusion}} & \multicolumn{2}{c}{PVN3D\Citep{pvn3d}} & \multicolumn{2}{c}{FFB6D\Citep{ffb6d}} & \multicolumn{2}{c}{Uni6D\Citep{uni6d}} & \multicolumn{2}{c}{Ours} \\ 
    \midrule
 Object & \small{ADD-S} & \small{ADD(S)} & \small{ADD-S} & \small{\small{ADD(S)}} & \small{ADD-S} & \small{ADD(S)} & \small{ADD-S} & \small{ADD(S)} & \small{ADD-S} &\small{ADD(S)} & \small{ADD-S} &\small{ADD(S)}\\ \midrule
        002\_master\_chef\_can & 83.9 & 50.2 & 95.3 & 70.7 & 96 & 80.5 & 96.3 & 80.6 & 95.4 & 70.2 & 96.0 & 74.2  \\
        003\_cracker\_box & 76.9 & 53.1 & 92.5 & 86.9 & 96.1 & 94.8 & 96.3 & 94.6 & 91.8 & 85.2 & 96.0 & 94.2 \\ 
        004\_sugar\_box & 84.2 & 68.4 & 95.1 & 90.8 & 97.4 & 96.3 & 97.6 & 96.6 & 96.4 & 94.5 & 97.6 & 96.6 \\ 
        005\_tomato\_soup\_can & 81.0 & 66.2 & 93.8 & 8.47 & 96.2 & 88.5 & 95.6 & 89.6 & 95.8 & 85.4 & 96.1 & 86.6 \\
        006\_mustard\_bottle & 90.4 & 81.0 & 95.8 & 90.9 & 97.5 & 96.2 & 97.8 & 97.0 & 95.4 & 91.7 & 97.8 & 96.7 \\   
        007\_tuna\_fish\_can & 88.0 & 70.7 & 95.7 & 79.6 & 96.0 & 89.3 & 96.8 & 88.9 & 95.2 & 79.0 & 96.3 & 76.0 \\ 
        008\_pudding\_box & 79.1 & 62.7 & 94.3 & 89.3 & 97.1 & 95.7 & 97.1 & 94.6 & 94.1 & 89.8 & 96.6 & 94.7 \\ 
        009\_gelatin\_box & 87.2 & 75.2 & 97.2 & 95.8 & 97.7 & 96.1 & 98.1 & 96.9 & 97.4 & 96.2 & 98.0 & 97.0 \\  
        010\_potted\_meat\_can & 78.5 & 59.5 & 89.3 & 79.6 & 93.3 & 88.6 & 94.7 & 88.1 & 93.0 & 89.6 & 95.7 & 91.9 \\ 
        011\_banana & 86.0 & 72.3 & 90.0 & 76.7 & 96.6 & 93.7 & 97.2 & 94.9 & 96.4 & 93.0 & 98.0 & 96.9 \\ 
        019\_pitcher\_base & 77.0 & 53.3 & 93.6 & 87.1 & 97.4 & 96.5 & 97.6 & 96.9 & 96.2 & 94.2 & 97.5 & 96.9 \\  
        021\_bleach\_cleanser & 71.6 & 50.3 & 94.4 & 87.5 & 96.0 & 93.2 & 96.8 & 94.8 & 95.2 & 91.1 & 97.0 & 95.3 \\  
        \textbf{024\_bowl} & 69.6 & 69.6 & 86.0 & 86.0 & 90.2 & 90.2 & 96.3 & 96.3 & 95.5 & 95.5 & 96.8 & 96.8 \\ 
        025\_mug & 78.2 & 58.5 & 95.3 & 83.8 & 97.6 & 95.4 & 97.3 & 94.2 & 96.6 & 93.0 & 97.7 & 96.3 \\  
        035\_power\_drill & 72.7 & 55.3 & 92.1 & 83.7 & 96.7 & 95.1 & 97.2 & 95.9 & 94.7 & 91.1 & 97.6 & 96.8 \\  
        \textbf{036\_wood\_block} & 64.3 & 64.3 & 89.5 & 89.5 & 90.4 & 90.4 & 92.6 & 92.6 & 94.3 & 94.3 & 96.1 & 96.1 \\ 
        037\_scissors & 56.9 & 35.8 & 90.1 & 77.4 & 96.7 & 92.7 & 97.7 & 95.7 & 87.6 & 79.6 & 95.0 & 90.3 \\  
        040\_large\_marker & 71.7 & 58.3 & 95.1 & 89.1 & 96.7 & 91.8 & 96.6 & 89.1 & 96.7 & 92.8 & 97.0 & 93.1 \\
        \textbf{051\_large\_clamp} & 50.2 & 50.2 & 71.5 & 71.5 & 93.6 & 93.6 & 96.8 & 96.8 & 95.9 & 95.9 & 97.0 & 97.0 \\  
        \textbf{052\_extra\_large\_clamp} & 44.1 & 44.1 & 70.2 & 70.2 & 88.4 & 88.4 & 96.0 & 96.0 & 95.8 & 95.8 & 96.5 & 96.5 \\  
        \textbf{061\_foam\_brick} & 88.0 & 88.0 & 92.2 & 92.2 & 96.8 & 96.8 & 97.3 & 97.3 & 96.1 & 96.1 & 97.4 & 97.4 \\ 
        \midrule
        Avg & 75.8 & 59.9 & 91.2 & 82.9 & 95.5 & 91.8 & 96.6 & 92.7 & 95.2 & 88.8 & 96.8 & 91.5 \\   \bottomrule
\end{tabular}}
\end{center}                        
\end{table*}

\begin{table*}[ht]
\begin{center}
\caption{Evaluation results (ACC \small{ADD(S)}-0.1d) on the LineMOD dataset. Symmetric objects are denoted in bold.}

\label{tab:LineMOD-ADD}
\resizebox{1\linewidth}{!}{
\begin{tabular}{lcccccc}
    \toprule
 & PoseCNN\Citep{posecnn} & DenseFusion\Citep{densefusion} & PVN3D\Citep{pvn3d} & FFB6D\Citep{ffb6d} & Uni6D\Citep{uni6d} & Ours \\
    \midrule
      ape & 77.0   & 92.3 & 97.3 & 98.4 & 93.71 & 95.71 \\
      benchvise & 97.5  & 93.2 &  99.7 & 100.0 & 99.81 & 99.90 \\
      camera & 93.5 & 94.4  & 99.6 & 99.9 & 95.98 & 95.78\\
      can & 96.5  & 93.1 & 99.5 & 99.8 & 99.02 & 96.01\\
      cat & 82.1  & 96.5 & 99.8 & 99.9 & 	98.10 & 99.20\\
      driller & 95.0  & 87.0 & 99.8 & 100.0 & 99.11 & 99.21\\
      duck & 77.7  & 92.3 & 97.7 & 98.4 & 89.95 & 92.11\\
      \textbf{eggbox} & 97.1  & 99.8 & 99.8 & 100.0 & 100.00 & 100.00 \\
      \textbf{glue} & 99.4  & 100.0 & 100.0 & 100.0 & 99.23 & 99.61\\ 
      holepuncher & 52.8 & 92.1 &  99.9 & 99.8 & 90.20 & 92.01\\
      iron & 98.3  & 97.0  & 99.7 & 99.9 & 99.49 & 97.96\\
      lamp & 97.5 & 95.3  & 99.8 & 99.9 & 99.42 & 98.46\\
      phone & 87.7  & 92.8  & 99.5 & 99.7 &97.41 & 97.69\\
      \hline
      Avg & 88.6 & 94.3  & 99.4 & 99.7 & 97.03  & 97.20\\   \bottomrule
\end{tabular}}
\end{center}                        
\end{table*}

\begin{table*}[t]
\begin{center}
\caption{Evaluation results (ACC \small{ADD(S)}-0.1d) on the Occlusion LineMOD dataset. Symmetric objects are denoted in bold.}
\vspace{.15in}
\label{tab:ocLineMOD-ACC}
\resizebox{0.99\linewidth}{!}{
\begin{tabular}{lccccc}
    \toprule
 Method & PoseCNN\Citep{posecnn}   & PVN3D\Citep{pvn3d} & FFB6D\Citep{ffb6d} & Uni6D\Citep{uni6d} & Ours \\
    \midrule
      ape & 9.6  &  33.9 & 47.2 & 32.99 & 44.26 \\
      can & 45.2  & 88.6 & 85.2 & 51.04 & 53.33\\
      cat & 0.9  & 39.1 & 45.7 & 4.56 & 16.70\\
      driller & 41.4 & 78.4 & 81.4 & 58.40 & 63.02\\
      duck & 19.6  & 41.9 & 53.9 & 34.80 & 38.09\\
      \textbf{eggbox} & 22.0  & 80.9 & 70.2 & 1.73 & 4.60  \\
      \textbf{glue} & 38.5   & 68.1 & 60.1 & 30.16 & 40.27 \\ 
      holepuncher & 22.1  & 74.7 & 85.9 & 32.07 & 50.93\\
       \hline
        Avg & 24.9  & 63.2 & 66.2 & 30.71 & 40.15 \\ \bottomrule
\end{tabular}}
\end{center}                        
\end{table*}

\subsection{Comparison with Other Methods}
We compare our method with others on YCB-Video, LineMOD, and Occlusion LineMOD datasets.

\subsubsection{Evaluation on YCB-Video Dataset}
We present the category-level results of the proposed Uni6Dv2 on YCB-Video dataset in Table~\ref{tab:ycb}. Compared with other methods, our approach advances state-of-the-art results by 0.2$\%$ on the ADD-S metric and achieves 91.5$\%$ on the ADD(S) metric. It is worth emphasizing that Uni6Dv2 comprehensively exceeds Uni6D, especially on objects with fewer pixels.
For example, "037\_scissors" of Uni6Dv2 achieves 7.36$\%$ improvement on the ADD-S metric and 9.72$\%$ on the ADD(S).

\subsubsection{Evaluation on LineMOD Dataset}
Experimental results of LineMOD dataset are reported in Table~\ref{tab:LineMOD-ADD}, our approach achieves 97.20\% ACC ADD(S)-0.1d. Compared with the original Uni6D, it improves the performance in "ape", "benchvise", "cat", "driller", "duck", "glue", "holepuncher" and "phone". Because the LineMOD dataset has less depth noise and occlusion than the YCB-Video dataset, our two-step denoising method make improvement slightly.

\subsubsection{Evaluation on Occlusion LineMOD Dataset}
We train our model on the LineMOD dataset and evaluate it on the Occlusion LineMOD dataset to obtain the performance on occlusion scene. Experimental results of Occlusion LineMOD are reported in Table~\ref{tab:ocLineMOD-ACC}. Compared with Uni6D, our method obtains 9.44\% improvement on the ACC ADD(S)-0.1d metric, and the ACC ADD(S)-0.1d of all categories have been improved.

\subsection{Time Efficiency}
\begin{figure}[t]
    \centering
    \includegraphics[width=0.45\textwidth]{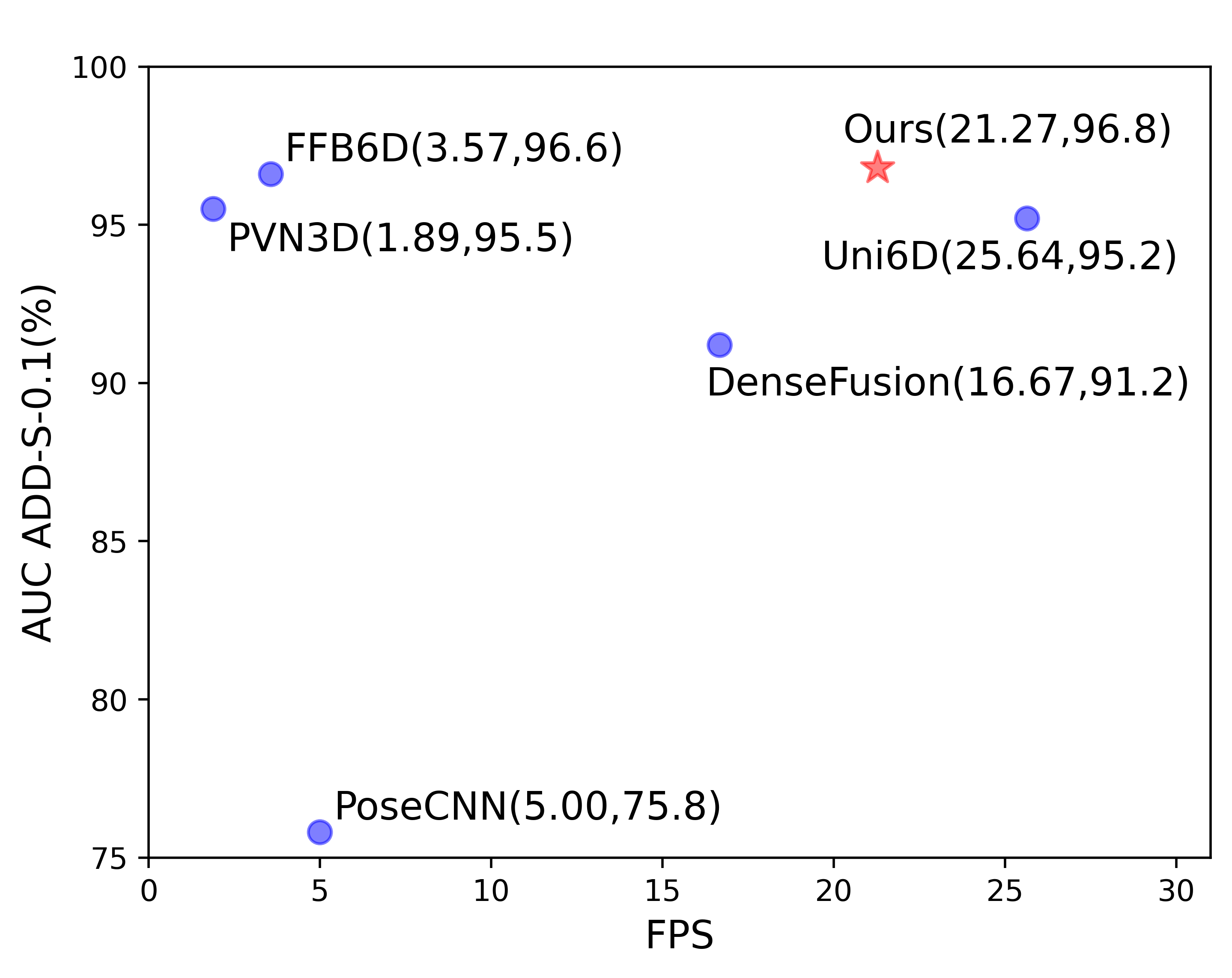}
    \caption{Comparison of effectiveness and efficiency.}
    \label{fig:efficieny}
\end{figure}

We compare the inference speed of our method with PoseCNN\Citep{posecnn}, DenseFusion\Citep{densefusion}, PVN3D\Citep{pvn3d}, FFB6D\Citep{ffb6d}, and Uni6D\Citep{uni6d} in Table~\ref{tab:time} and Fig.~\ref{fig:efficieny}.
Our method achieves 21.27 FPS which is $6\times$ faster than FFB6D (SOTA of keypoint-based methods). Compared with Uni6D (SOTA of RoI-based methods), our methods only sacrifices 17\% efficiency to achieve significant gains, especially without real training data in Table~\ref{tab:lack-real} (\textbf{+20.11\%} ADD-S and \textbf{+29.48\%} ADD(S)). 
%
More specifically, the first and second steps of our method take \SI{35}{ms} and \SI{12}{ms}, respectively. The extra computation primarily concentrated in the backbone of the second step, causing our method to perform somewhat slower than Uni6D.

\begin{table}[htb]
   \centering
     \caption{Time cost and frames per second (FPS) on YCB-Video Dataset.
     } 
     \vspace{.15in}
     \resizebox{1.0\linewidth}{!}{
    \begin{tabular}{ccccc}
    \toprule
        Method & Network/ms & Post-process/ms & All/ms & FPS \\
        \midrule
        PoseCNN\Citep{posecnn} & 200 & \textbf{0} & 200 & 5 \\
        DenseFusion\Citep{densefusion} & 50 & 10 & 60 &  16.67\\
        PVN3D\Citep{pvn3d} & 110 & 420 & 530 & 1.89\\
        FFB6D\Citep{ffb6d} & \textbf{20} & 260 & 280 & 3.57 \\
        Uni6D\Citep{uni6d} & 39 & \textbf{0} & \textbf{39} & \textbf{25.64} \\
        Ours & 47 & \textbf{0} & 47 & 21.27 \\
    \bottomrule
    \end{tabular}}
    \label{tab:time} 
\end{table}

\makeatletter
\newcommand\figcaption{\def\@captype{figure}\caption}
\newcommand\tabcaption{\def\@captype{table}\caption}
\makeatother

%% file: sec/5exp_ablation.tex
\subsection{Ablation Study}

\subsubsection{Effect of Noise Elimination}

To investigate the contribution of 
the proposed noise elimination operations, we superimpose these denoising operations gradually.
In this comparison, we divide instance-outside noise into noise within and out of the detection bounding box, and eliminate them by cropping and masking. 
Uni6D without any denoising operation is our baseline. In Table~\ref{tab:noise-elimination-v2}, "Box" denotes cropping the input data by detection bounding boxes to eliminate noise in the background out of the box. "Mask" means filtering out the background noise in the box, and "Depth" denotes calibrating the depth feature with the depth denoising module. "Box" and "Mask" are used together for instance-outside noise elimination, and "Depth" is used for instance-inside noise elimination.
\begin{table}[ht]
  \caption{Comparing different types of denoising on the YCB-Video dataset.}
  \vspace{.15in}
  \label{tab:noise-elimination-v2}
  \centering
   \resizebox{0.9\linewidth}{!}{
    \begin{tabular}{ccccccc}
        \toprule
      \multicolumn{3}{c}{Denoising level}  & \multicolumn{2}{c}{Based on GT} & \multicolumn{2}{c}{Based on DT} \\ 
        \midrule
     \small{Box} & \small{Mask} & \small{Depth} & \small{\small{ADD(S)}} & \small{ADD-S} & \small{ADD(S)} & \small{ADD-S}\\ \midrule[1pt]
                &  &  & - & - & 95.18 & 88.83 \\
              \checkmark &  &  & 96.39 & 90.58 & 96.01 & 89.80 \\
              \checkmark& \checkmark &  & 96.85 & 91.55& 96.52 & 91.27  \\ 
            \checkmark & \checkmark & \checkmark& \textbf{97.67} & \textbf{92.52} &\textbf{96.77} & \textbf{91.51}  \\ 
      \bottomrule
    \end{tabular}}
\vspace{-0.25cm}
    
\end{table}
\begin{table}[ht]
      \caption{Comparing different denoising strategies for instance-outside noise on the YCB-Video dataset.}
      \vspace{.15in}
      \label{tab:mask-design}
      \centering
      \resizebox{0.8\linewidth}{!}{
        \begin{tabular}{ccc}
          \toprule
            Denoising strategy & ADD-S & ADD(S) \\
            \midrule
            no denoising & 95.18 & 88.83\\
            feature-level denoising & 96.25 & 90.21\\
            image-level denoising & \textbf{96.52} & \textbf{91.27} \\
        \bottomrule
      \end{tabular}}
\end{table}
"Based on GT" means using the ground truth of detection, segmentation and reprojection depth to train the pose regression model, it indicates that eliminating every noise increases the upper bound on performance.
"Based on DT" represents the performance of our method without any ground truth leaking. We observe that eliminating every type of noise leads to an improvement in performance. Specifically, using the first denoising step to remove instance-outside noise leads to a 1.34$\%$ and 1.44$\%$ improvement on ADD-S and ADD(S), respectively,  while the second denoising step further improves performance by 0.25$\%$ on ADD-S and 0.24$\%$ on ADD(S).

\subsubsection{Comparison of Denoising Strategies}
To further investigate the effect of our two-step denoising method, we compare various strategies for removing instance-outside and instance-inside noise. 

 For instance-outside noise elimination, there are two optional strategies: feature-level and image-level denoising. The feature-level denoising crops and masks the feature map directly, which can reduce computational complexity. However, it does not completely eliminate the noise from the background around the instance introduced by the receptive field, which limits the denoising performance. As shown in Table~\ref{tab:mask-design}, we adopt the image-level denoising in the first step, which brings 0.27$\%$ improvement on ADD-S and 1.06$\%$ on ADD(S) compared to feature-level denoising.
 Another advantage of using image-level denoising strategy is that it allows us to avoid introducing UV information in the input of instance segmentation network. The UV information can break the convolution translation invariance and negatively affects the quality of detection and segmentation. To evaluate the effect of UV input for Mask-RCNN, we compare the performance of using RGB-D and RGB-D with UV input for Mask-RCNN. As shown in Table~\ref{tab:mask-quality}, introducing UV in Mask R-CNN reduces the performance and using RGB-D image as the input data is better.

 For instance-inside noise elimination, we explore the effect of adding different depth-related information as the supervised labels, including the re-projected depth, XY and NRM. Results in Table~\ref{tab:calibration-design} demonstrate that using all depth-related information strengthen the effect of denoising, which brings 0.25$\%$ improvement on ADD-S and 0.28$\%$ on ADD(S) for the YCB-Video dataset.

\begin{table}[htb]
        \caption{Comparing different inputs of Mask-RCNN on the YCB-Video dataset.}
        \vspace{.15in}
      \label{tab:mask-quality}
      \centering
       \resizebox{0.9\linewidth}{!}{
        \begin{tabular}{cccc}
          \toprule
            Input of Mask-RCNN  & mIoU & ADD-S & ADD(S) \\
            \midrule
            RGB-D with UV & 78.15 & 96.52 & 91.28\\
            RGB-D & 80.43 & 96.77 & 91.51 \\
        \bottomrule      \end{tabular}}
\end{table}

\begin{table}[htb]
   \vspace{-0.6cm}
    \caption{Comparison of instance-inside denoise strategies on the YCB-Video dataset.}
    \vspace{.15in}
      \label{tab:calibration-design}
      \centering
      \resizebox{0.99\linewidth}{!}{
        \begin{tabular}{cccccc|cc}
        \toprule
        Depth Denoise & XY Denoise & NRM Denoise & ADD-S & ADD(S) \\
        \midrule
        &  &  & 96.52 & 91.27 \\
        \checkmark &  &  & 96.51 & 90.30\\
        \checkmark & \checkmark &  &  96.74 & 91.26 \\
        \checkmark &  & \checkmark &  96.74 & 91.23 \\
        \checkmark & \checkmark &  \checkmark & \textbf{96.77} & \textbf{91.51} \\
        \bottomrule
      \end{tabular}}

\end{table}

\subsubsection{Advantage of Reducing Real Data Requirements}

 To verify the advantage of two denoising strategies in reducing real data requirements, We randomly sample 0$\%$, 1$\%$ and 10$\%$ of the available real training data and fused it with synthetic data to create new training datasets, the test dataset keeps the original setting.
%
%
%
 The gains from removing instance-outside noise and instance-inside noise are shown in Table~\ref{tab:lack-real}, where "0\%" indicates training models only with synthetic data. 
Since all synthetic depth images are randomly masked in order to simulate holes in the realistic depth images, it is required to eliminate instance-inside noise even when training only with synthetic data.
Compared with the baseline (Uni6D), individually removing the two types of noise achieves significant improvements and removing both achieves SOTA.
Especially when training without real data, removing instance-outside noise gains 19.96\% on ADD-S and 29.27\% on ADD(S), and removing instance-inside noise gains 19.38\%  on ADD-S and 29.08\% on ADD(S).

\begin{table}[tbp]
\begin{center}
\caption{The superiority of the proposed method in reducing real data requirements on the YCB-Video dataset.}
\vspace{.15in}
     \label{tab:lack-real} 
     \resizebox{0.99\linewidth}{!}{
        \begin{tabular}{cccccc}
            \toprule
        \multicolumn{2}{c}{Denoising}  & \multicolumn{4}{c}{Percentage of real training data} \\
            \midrule
          outside & inside & 100\%  & 10\%   & 1\%   & 0\%  \\\midrule[1pt]
                 &  & 95.18/88.83 & 91.84/81.47 &  89.95/77.98 & 73.52/54.71  \\
                \checkmark &  & 96.52/91.27 & 95.48/87.59 & 93.73/84.37 & 93.48/83.98  \\
                &  \checkmark & 95.77/89.75 & 94.14/85.80 & 93.49/83.99 & 92.90/83.79 \\
                \checkmark& \checkmark & \textbf{96.77/91.51} & \textbf{95.60/88.02} & \textbf{95.22/86.41} & \textbf{93.63/84.19} \\\bottomrule
        \end{tabular}}
\end{center}                        
\end{table}

\begin{figure*}[htbp]

    \centering
    \includegraphics[width=0.8\textwidth]{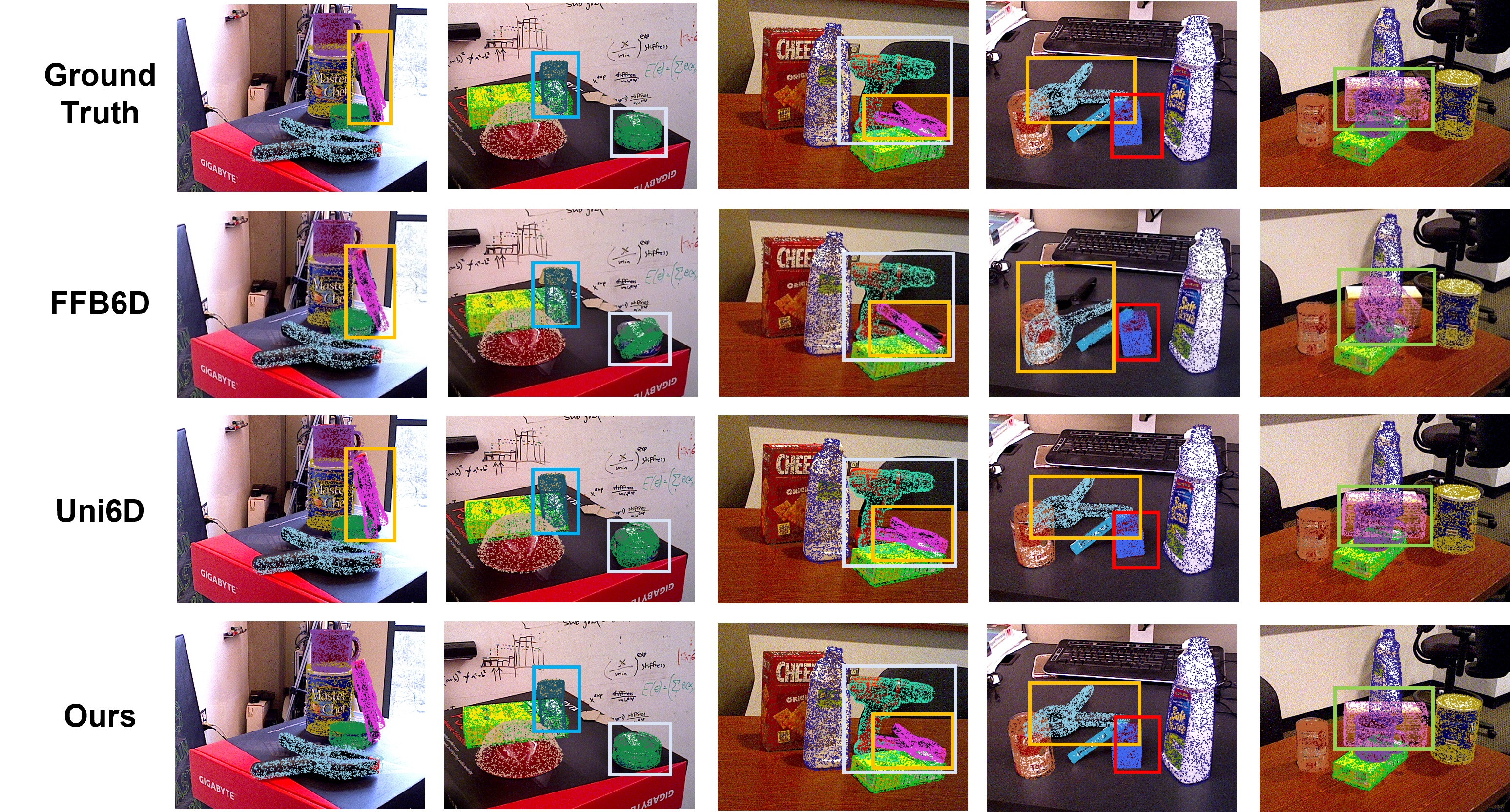}
    \caption{Qualitative results of 6D pose on the YCB-Video dataset.}
    \label{fig:supp_ycb}
\end{figure*}

\subsection{Qualitative Results}

For intuitive comparison, the qualitative results of Uni6Dv2 and the baseline Uni6D\Citep{uni6d} as well as the SOTA FFB6D\Citep{ffb6d} on the YCB-Video dataset are shown in Fig.~\ref{fig:supp_ycb}, with failed results framed by the bounding box. 
Uni6Dv2 achieves a more precise and robust result when there are more noise interference.
More qualitative results on the LineMOD dataset are provided in our supplementary materials.

\vspace{-0.4cm}
\subsection{Denoising on Other RoI-based Methods}
We take the latest RoI-based method ES6D\Citep{es6d}  as an example, which feeds cropped RGB images and masked 3D coordinates maps into 6D pose estimator. With the background depth removed, ES6D only suffers from the instance-inside noise from depth sensors. Thus, we apply our denoising method on ES6D to remove its instance-inside noise, and the results are provided in Table~\ref{tab:noise-elimination-v2-es6d}. We observe that removing instance-inside noise by depth denoising module gains 0.5\% on ADD-S and 1.0\% on ADD(S).

\begin{table}[tbp]
\begin{center}
  \caption{Comparing different types of denoising for ES6D on the YCB-Video dataset.}
  \vspace{.15in}
  \label{tab:noise-elimination-v2-es6d}
   \resizebox{0.65\linewidth}{!}{
    \begin{tabular}{ccc}
        \toprule
     \multicolumn{1}{c}{Denoising}  & \multicolumn{2}{c}{Metric} \\ 
        \midrule
     inside & ADD-S & ADD(S)\\ \midrule[1pt]
                & 93.1 & 89.0  \\
                \checkmark & \textbf{93.6(+0.5)} & \textbf{90.0(+1.0)} \\
      \bottomrule
    \end{tabular}}
\end{center}                        
\end{table}

\subsection{Implementation Details}
\label{sec:implementation_details}

We adopt ResNet-50\Citep{he2016deep} and FPN\Citep{lin2017feature} for Mask R-CNN. The first convolutional layer's channels are set to 4 for the input RGB-D image. We trained the models using 16 NVIDIA GTX 1080Ti GPUs, with three images per GPU, for 40 epochs, using an SGD optimizer with a momentum of 0.9 and weight-decay of 0.0001. The initial learning rate is set to 0.005 with a linear warm-up. and decreased by 0.1 after 15, 25 and 35 epochs. $\lambda_0$ in the loss function is set to 1 in epoch [1, 20), 5 in [20, 30), 20 in [30, 38) and 50 in [38, 40],  $\lambda_1$ and $\lambda_2$ are set to 1. Please refer to our supplementary materials for more details.



%% file: sec/6conclusion.tex
\section{CONCLUSION}
\label{sec:conclusion}
In this paper, we improve Uni6D from the perspective of denoising. We first investigate the instance-outside and instance-inside noise which have severely limited the original Uni6D's performance. To address this issue, we present Uni6Dv2, which employs a two-step denoising pipeline. In the first step, an instance segmentation network is used to filter out instance-outside noise, and in the second step, a lightweight depth denoising module is used to correct instance-inside noise.
Extensive experimental results show that our method outperforms other state-of-the-art methods and improves the performance of the original Uni6D with only a slight decrease in inference efficiency. 

As limitations, more elegant denoising methods beyond cropping and masking from RGB-D data should be investigated. Furthermore, our method still suffers from the translation distribution gap between training and testing datasets (the model is hard to generalize to different 3D positions), which is a common issue for direct regression methods. More efficient and effective innovative methods need to be explored to further fix this issue. Despite these limitations, we believe that Uni6Dv2 can be a strong baseline approach to 6D pose estimation and inspires future work.

\section*{Acknowledgments}
This work was also sponsored by Hetao Shenzhen-Hong Kong Science and Technology Innovation Cooperation Zone: HZQB-KCZYZ-2021045.

%% file: sec/supplement.tex
%
%




%

%

\onecolumn
\aistatstitle{Supplementary Materials}
\section{Implementation Details}
In this section, we first provide a detailed description of the re-projected depth, XY, and NRM in denoising module. Then we describe the details of training on YCB-Video\Citep{calli2015ycb} and LineMOD\Citep{hinterstoisser2011multimodal}. 



\subsection{The details of the training phase}
We provide the details of training on YCB-Video and LineMOD in Table~\ref{tab:details_ycb} and Table~\ref{tab:details_lm}. 
In each step, we use the ImageNet pre-trained weights to initialize the backbone except the first convolutional layer. 
The first convolutional layer is initialized by kaiming uniform because the number of channels is changed to accommodate to the new input. 
For some categories in the LineMOD dataset, including "cat", "eggbox" and "glue", the mask operation is not applied to the position embedding (PE).

\begin{table}[htbp]
  \centering
  \caption{The details of the training on YCB-Video.}
      \label{tab:details_ycb}
    \begin{tabular}{c|p{14.055em}|p{14.5em}}
    \toprule
          & The first step & The second step \\
    \midrule
    Input data & RGB-D & RGB-D, UV, PE, XY, NRM \\ 
    \midrule
    \multirow{11}{*}{Augmentation} & \textbf{Multi-scale training}: [320, 400, 480, 600, 720] (max size is 900); & \textbf{Multi-scale training}: [180, 200, 224, 250, 270] (max size is 360); \\
          & \textbf{Background replacing}: replace the background of the synthetic data with the real image background; & \textbf{Background replacing}: replace the background of the synthetic data with the real image background; \\
          & \textbf{Random crop}: 0.3 probability, and keep all objects; & \textbf{Random crop}: 1.0 probability, expand detection box by 0.3 and keep the object intact; \\
          &  & \textbf{Mask dilation and erotion}: 0.75 probability, the size of kernel random from 3, 5 and 7. \\
    \midrule
    {\multirow{7}{*}{Training }} & \textbf{Pretrained weight}: ImageNet; & \textbf{Pretrained weight}: ImageNet; \\
          & \textbf{Schedule}: 40epoch, MultiStepLR with [15, 25, 35] schedule and 0.1$\times$decay; & \textbf{Schedule}: 40epoch, MultiStepLR with [15, 25, 35] schedule and 0.1$\times$decay; \\
          & \textbf{Optimizer}: SGD, momentum 0.9, weight\_deacy 0.0001, warm-up 4 epoch. & \textbf{Optimizer}: SGD, momentum 0.9, weight\_deacy 0.0001, warm-up 4 epoch. \\
    \midrule
    {\multirow{4}{*}{Loss function}} & $\mathcal{L} = \mathcal{L}_{mask} 
    +   \mathcal{L}_{bbox} +\mathcal{L}_{cls} +\mathcal{L}_{rpn} $ & $\mathcal{L} = \lambda_0 \cdot \mathcal{L}_{rt} +  \mathcal{L}_{abc} + \mathcal{L}_{depth}$ \\
          &  & 
          $\lambda_0$ is changed in training: 1-15 epoch is 1, 16-25 epoch is 5, 26-35 epoch is 10 and 36-40 epoch is 20.  \\
    \bottomrule
    \end{tabular}%
\end{table}%

\begin{table}[htbp]
  \centering
  \caption{The details of the training on LineMOD.}
  \vspace{.15in}
      \label{tab:details_lm}
    \begin{tabular}{l|p{14.055em}|p{14.5em}}
    \toprule
          & The first step & The second step \\
    \midrule
    \multicolumn{1}{p{7.665em}|}{Input data} & RGB-D & RGB-D, UV, PE, XY, NRM \\ 
    \midrule
    \multirow{8}{*}{Augmentation} & \textbf{Multi-scale training}: [320, 400, 480, 600, 720] (max size is 900); & \textbf{Multi-scale training}: [180, 200, 224, 250, 270] (max size is 360); \\
          & \textbf{Background replacing}: replace the background of the synthetic data with the real image background; & \textbf{Background replacing}: replace the background of the synthetic data with the real image background; \\
          & \textbf{Random crop}: 0.3 probability, and keep all objects. & \textbf{Random crop}: 1.0 probability, expand detection box by 0.3 and keep target. \\
    \midrule
    \multirow{7}{*}{Training } & \textbf{Pretrained weight}: ImageNet; & \textbf{Pretrained weight}: ImageNet; \\
          & \textbf{Schedule}: 40epoch, MultiStepLR with [15, 25, 35] schedule and 0.1$\times$decay; & \textbf{Schedule}: 40epoch, MultiStepLR with [15, 25, 35] schedule and 0.1$\times$decay; \\
          & \textbf{Optimizer}: SGD, momentum 0.9, weight\_deacy 0.0001, warm-up 4 epoch. & \textbf{Optimizer}: SGD, momentum 0.9, weight\_deacy 0.0001, warm-up 4 epoch. \\
    \midrule
    \multirow{4}{*}{Loss function} & $\mathcal{L} = \mathcal{L}_{mask} 
    +   \mathcal{L}_{bbox} +\mathcal{L}_{cls} +\mathcal{L}_{rpn} $ & $\mathcal{L} = \lambda_0 \cdot \mathcal{L}_{rt} +  \mathcal{L}_{abc} + \mathcal{L}_{depth}$ \\
          &  & 
          $\lambda_0$ is changed in training: 1-15 epoch is 1, 16-25 epoch is 5, 26-35 epoch is 10 and 36-40 epoch is 20.  \\
    \bottomrule
    \end{tabular}%
\end{table}%

\section{More Qualitative Results}
\begin{figure*}[htbp]
    \centering
    \includegraphics[width=0.55\textwidth]{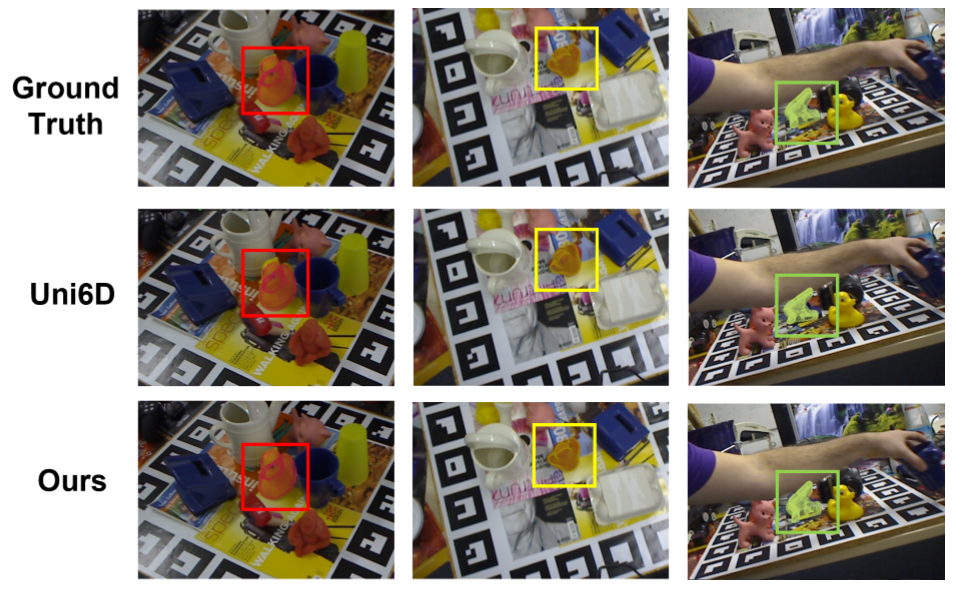}
    \caption{Qualitative results of 6D pose on the LineMOD dataset.}
    \label{fig:supp_lm}
\end{figure*}

We provide more qualitative comparison results between our method and the original Uni6D on LineMOD, which are shown in Fig.~\ref{fig:supp_lm}. In addition, we recommend readers to watch the video from \href{https://youtu.be/dhVq1uqSZoE}{https://youtu.be/dhVq1uqSZoE}, which shows a more comprehensive comparison between our method and the Uni6D. Compared with Uni6D, our method achieves more precise and robust performance.

